\documentclass[conference]{IEEEtran}
\usepackage[utf8]{inputenc} 
\usepackage{cite}
\usepackage{booktabs}
\newcommand{\argmax}{\operatornamewithlimits{argmax}}
\usepackage{graphicx}
\usepackage{booktabs}
\usepackage{multirow}

\ifCLASSINFOpdf
\else
\fi
\usepackage{amsmath}
\usepackage{algpseudocode}
\usepackage[ruled,norelsize]{algorithm2e}
\begin{document}
%
\title{Applying Naïve Bayes Classification to Google Play Apps Categorization}

\author{\IEEEauthorblockN{Babatunde Olabenjo}
\IEEEauthorblockA{Department of Computer Science\\
University of Saskatchewan\\
Saskatoon, SK, Canada}}


%


\maketitle

\begin{abstract}
There are over one million apps on Google Play Store and over half a million publishers. Having such a huge number of apps and developers can pose a challenge to app users and new publishers on the store. Discovering apps can be challenging if apps are not correctly published in the right category, and, in turn, reduce earnings for app developers. Additionally, with over 41 categories on Google Play Store, deciding on the right category to publish an app can be challenging for developers due to the number of categories they have to choose from. Machine Learning has been very useful, especially in classification problems such sentiment analysis, document classification and spam detection. These strategies can also be applied to app categorization on Google Play Store to suggest appropriate categories for app publishers using details from their application.

In this project, we built two variations of the Naïve Bayes classifier using open metadata from top developer apps on Google Play Store in other to classify new apps on the store. These classifiers are then evaluated using various evaluation methods and their results compared against each other.  The results show that the Naïve Bayes algorithm performs well for our classification problem and can potentially automate app categorization for Android app publishers on Google Play Store.
\end{abstract}


%
\IEEEpeerreviewmaketitle

\section{Introduction}
Machine Learning has been widely used in various domains to study and learn from patterns in data to make accurate predictions. The use of machine learning can be seen in our daily lives, especially in email providers for detecting spam messages.  With the increase in data in recent years, understanding and making appropriate decisions can be challenging due to the vast amount of data to be analyzed. Furthermore, because the data is in different forms, making accurate decisions can be overwhelming. Machine Learning can allow us to make proper decisions from this vast amount of data by allowing the computer to learn statistically from the large data set and make predictions for new instances based on what has been learnt previously. Also, machines can be trained to learn by feeding it with examples where it then makes decisions based on the carefully selected examples provided.

To understand how machine learning systems work, we can classify them into three broad categories based on the nature of the learning system. These learning systems as described by \cite{Russell2009} are:\\

\begin{enumerate}

\item {\textit{Supervised Learning:} In supervised learning, the machine or computer is given a set of examples to learn from. The machine learns from the inputs (examples) and makes predictions based on the examples provided.}
\item{\textit{Unsupervised Learning:} In unsupervised learning, the machine is left to learn from the data provided to it in order to discover patterns that can be used to make predictions eventually.}
\item{\textit{Reinforcement Learning:} Reinforcement learning, for example, a computer learning to play a game, is the process whereby the computer learns in a dynamic environment to perform a certain task without explicitly being told if it is close to achieving the goal. This way, the computer learns from the mistakes it has previously made and from the reward it gets from achieving a particular goal \cite{Russell2009}.\\}

\end{enumerate}

In recent years, mobile phones have proven useful in our daily lives especially with the increased availability and reduction in cost. Furthermore, with the advent of App Stores for hosting mobile applications thus providing a variety of tools useful in our daily lives, these mobile devices become more and more integrated into our lives. Google Play Store is one of the biggest App Stores with millions of applications and the official App Store for hosting Android applications for the Android operating system. Finding useful applications can be challenging to app downloaders because many applications are being placed in the wrong category by developers. At the same time, this affects the number of downloads an app will receive ultimately affecting the earnings of app publishers.

\subsection{Motivation}
Making accurate predictions for developers about what category an app should be uploaded to on Google Play Store will potentially improve the discovery of their applications and revenue at the long run. Classification of apps on the store is a useful application of machine learning. There is an increasing number of research using machine learning to classify text, sentiment analysis and documents with Naïve Bayes such as \cite{Eyheramendy2003,Frank2006,Howedi2014,Ting2011}. With the increasing amount of data available online, providing useful information from this data creates new knowledge and improves the overall success of businesses \cite{Ting2011}. Since most machine learning classification algorithms are time-consuming and complicated, using Naïve Bayes classification provides a fast and simple way to classify data \cite{Narayanan2013}.

Android app users visiting Google Play Store often find top apps listed whenever they search for an app in the store than non-top apps that may be useful to them. In some situations, these users are willing to look into app categories for apps related to a category they want. Looking for an app via categories in the store can become frustrating for users if there are many wrongly categorised apps; this can be a problem for developers as well, as app users will find it difficult to discover their apps. For example, The Sun Daily \footnote{TheSunDaily: http://www.thesundaily.my/news/871252} reported a case on Health apps where more than 50\% of apps were miscategorized leading to fewer downloads.

Deciding what category to upload an app to can be challenging for developers and Machine learning can be used to suggest suitable categories to developers based on the details they provide. In this research, we find out how machine learning, using supervised learning can suggest appropriate categories with data extracted from successful developers on Google Play Store. The success of an app on the store varies based on the description of the app, the category, whether the app is free or not including other factors. Using data from successful developers on Google Play Store, we can provide a supervised learning training set for efficient classification of apps on the store.

The aim of this research is to categorize apps on Google Play Store, using existing data from apps developed by top publishers on the store in order to suggest the best category for a new app. We use Term Frequency–Inverse Document Frequency (TF-IDF) statistics to extract useful information that can be used to build our classifier with Naive Bayes. The effectiveness of the classifier is measured using various validation methods, and the results are presented using a confusion matrix, f1-scores and other statistical variances. These validation methods include the k-fold cross-validation \cite{Fushiki2011}, shuffle-split cross-validation used to generate a learning curve that determines the training and test scores for various training data size \cite{Huh2001} and the recursive feature elimination for testing the number of features that produces the best results \cite{Granitto2006}.

\section{Literature Review}
There have been numerous application of machine learning in the industry; Amazon store, IBM e-commerce and others have employed machine learning in product classification as well as product recommendation. Advert placement and ad content design have been improved greatly by Google with machine learning. Machine learning has also been used extensively in image processing by Google for their image search. Although there are several machine learning algorithms available for various tasks, classification problems have become most predominant in this space. As described by \cite{Alpaydin2010} classification is an example of supervised learning, where a training set of observations correctly identified are fed into the machine learning algorithm to train the system. The process allows the machine learning algorithm to identify correctly new data provided, based on the knowledge acquired from the training set. In unsupervised learning, this process is known as clustering, and it involves data being grouped into categories based on some measure of similarity in the data.

Numerous research has been done to improve classification in various domain; an example is a research done by Schnack et al. \cite{Schnack2014} using machine learning to classify patients with schizophrenia, bipolar disorder and healthy subjects with their structural MRI scans. In their research, they used the Support Vector Machine (SVM) machine learning algorithm to create models from gray matter density images. There has also been similar research in product classification using SVM, for example \cite{Kreyenhagen2014} shows that SVM adds value to the classification of fashion brands in their research thereby making it easy for users to narrow down their searches when looking for a particular product. Other research such as sentiment analysis done by \cite{Nithya2014} used Naïve Bayes algorithm to classify the most identified features in an unstructured review and determine polarity distribution in terms of being positive, negative and neutral. Although little research has been published using Naïve Bayes for product classification, there are so many other classification problems in which the Naïve Bayes algorithm has been very effective.

\subsection{Applying Naïve Bayes to Classification Problems}
The Naive Bayes machine learning model \cite{Maron1961} is a popular statistical learning system that has been successful in many applications where features are independent of each other. An example of this model is found in the bag-of-words representation of text where the ordering of words is ignored. One of the earliest application of this model to Information Extraction was done by \cite{Freitag2000}. Information extraction involving extracting specific facts from text has also played a massive role in simplifying large dataset for users to understand. Zhenmei et al. \cite{Gu2006} proposed a smoothing strategy using Naïve Bayes for Information Extraction. The authors show that a well designed smoothing method will improve the performance of a Naïve Bayes Information Extraction learning system. \cite{Al-Aidaroo2012} compared Naïve Bayes with other classification algorithms for a medical dataset. Their results show that Naïve Bayes performed better than other algorithms in classifying medical datasets and can be applied to medical data mining; this is due to its simplicity and computational speed. Although Naïve Bayes have been criticized for its independence assumptions, the combination of Naïve Bayes and other classification algorithms can eventually improve its overall performance.

Other variations of Naïve Bayes algorithm have been proposed to improve the performance of the algorithm for various purposes. An example of this is the modified Naïve Bayes algorithm proposed by \cite{Thakur2014} to improve the classification of Nepali texts. Since the Nepali text is non-English and lacks basic linguistic components such as the stop words list, the stemmer, which involves removing morphological and inflexional endings from English words, and the Part-Of-Speech Tagger; the authors improved the performance of the classifier using lexicon domain pooling, and because the algorithm is flexible, it can be extended to other non-English languages like the Chinese or Japanese language. Another example of improving the Naive Bayes algorithm is the improved Naive Bayes probabilistic model-Multinomial Event Model for text classification by \cite{Qiang2010}. The model works by pushing down larger counts of word frequency because the Multinomial Naïve Bayes treats the occurrence of a word in a document independently even though multiple occurrences of the same word in a document are not necessarily independent \cite{Qiang2010} and Multinomial Naïve Bayes does not account for this occurrence. Other weighting schemes in text classification involve the use of N-grams, which is a sequence of n-items from a given document or text and (Term Frequency-Inverse Document Frequency) TF-IDF, which shows how important a word is in a document or a given set of documents \cite{Eck2005}.

One major application of the Naïve Bayes algorithm is in spam detection. \cite{Zhen2006} proposed a Naïve Bayes spam detection method based on decision trees, they also presented an improved method based on classifier error weight. Their experimentation shows that the implementation is valid, but there are not many solutions for valid incremental decision tree induction algorithm as they described \cite{Zhen2006}. Another use of Naïve Bayes is that proposed by \cite{Zhang2004} for ranking. The authors used a weighed Naïve Bayes for ranking in which each attribute has a different weight. Their results show that the weighted Naïve Bayes outperforms the standard Naïve Bayes, and both the weighted Naïve Bayes and the standard Naïve Bayes are better in performance than the decision tree algorithm \cite{Zhang2004}.

There has been a few research done on Google Play Store using sentiment analysis applied to customer reviews on mobile apps to determine their polarity such as \cite{Maalej2015}, where app reviews were automatically classified and result compared with other classification methods. From their results, the authors show that natural language processing with metadata from apps can improve classification precision significantly. This system can improve the design of review analytics tools for developers and app users when a large number of reviews is involved. Additionally, \cite{Islam2014} used sentiment analysis on customer reviews on the store. Although the authors did not use Naïve Bayes for analyzing the reviews, research such as \cite{Qiang2010, Zhen2006} and \cite{Maalej2015} show that machine learning using Naïve Bayes can also be applied in classifying customer reviews on Google Play Store. Most research show that machine learning using the Naïve Bayes algorithm can be implemented in numerous domains for classification problems. On Google Play Store, various research focus on sentiment analysis in customer reviews.  Little research has been done using machine learning to suggest appropriate categories for app developers with well known categorized applications on the store or to detect spam apps based on app metadata on the app store.

\section{Algorithm Description}
Naïve Bayes classifier used in this research is a simplified probabilistic classifier that is based on the Bayes theorem. Bayes theorem describes the probability of an event based on the conditions relating to the event. Bayes rule is defined mathematically as (\ref{eq:eq1})

\begin{equation}\label{eq:eq1}
P(A|B) = \frac{P(B|A) P(A)}{P(B)}
\end{equation}

where $A$ and $B$ are two events such that $P(A)$ which is the prior probability, and $P(B)$ are the probabilities of $A$ and $B$ independent of each other \footnote{Bayes Theorem: https://en.wikipedia.org/wiki/Bayes\_theorem}.\\

$P(A|B)$, is the posterior probability described as the conditional probability of observing event $A$ given that $B$ is true.\\

$P(B|A)$, is the likelihood described as the probability of observing event $B$ given that $A$ is true.\\

As discussed earlier in previous sections, the Naïve Bayes classifier has been used in various applications such as document classification, email spam detection and sentiment analysis. The classifier is based on an assumption that all attributes are independent of each other. Although this assumption makes other advanced classifier perform better in some scenarios, the Naïve Bayes classifier is known for its speed and less training set required to solve a classification problem. Since the classification of apps will be done on a regular computer, Naïve Bayes is most efficient in terms of CPU and memory consumption as described in \cite{Huang2003}.

\subsection{Theoretical Background}
From \cite{Manning2008} we can see that even though the probability estimates of Naïve Bayes is  sometimes of low quality, the classification decisions can provide good results. In text classification as described by \cite{Manning2008}, words are represented as tokens and classified into a particular class and by using the Maximum a Posterior (MAP) we can generate the classifier as described by \cite{Manning2008} as  (\ref{eq:eq2}).

\begin{align}\label{eq:eq2}
c_{MAP} &= \argmax_{c\in C} \left( P(c|d) \right) \nonumber\\
&= \argmax_{c\in C} \left( P(c) \prod_{1 \leq k \leq n_d} P(t_k|c)\right)
\end{align}

where $P(c|d)$ represents  conditional probability of class $c$ given document $d$\\  

$t_k$ represent the tokens of the document, $C$ represents the set of classes used in the classification.\\

$P(c)$ represents the prior probability of class $c$ and $P(t_k|c)$ represents the conditional probability of token $t_k$ given the class $c$.\\

Here, we estimate the likelihood, multiplied by the probability of a particular class prior for each class and select the class with the highest probability represented as $c_{map}$. In order to prevent underflow when calculating the product of the probabilities, we maximize the sum of their logarithms as described by \cite{Manning2008} using (\ref{eq:eq3}), thereby choosing classes with the highest log score represented as $c_{map}$.

\begin{equation}\label{eq:eq3}
c_{map} = \argmax_{c\in C} \left( \log P(c) + \sum_{1 \leq k \leq n_d} \log P(t_k|c)\right)
\end{equation}

Furthermore, if a word does not occur in a particular class, then the conditional probability is $0$ giving us $\log(0)$ which will eventually throw an error. To resolve this, we use Laplace smoothing by adding $1$ to each count, giving us (\ref{eq:eq4}).

\begin{align}\label{eq:eq4}
P(t|c) &= \frac{T_{ct}+1}{\sum\limits_{t^\prime \in V}(T_{ct^\prime}+1)} \nonumber\\
&= \frac{T_{ct}+1}{\left(\sum\limits_{t^\prime \in V}T_{ct^\prime}\right)+B^\prime}
\end{align}

\subsection{Application in Google Play Store App Categorization}
To build a Naïve Bayes classifier that would classify words for our dataset as features for a particular category, we would use two variations; the Multinomial and Bernoulli Naïve Bayes classifiers. The Multinomial Naïve Bayes classifier is used when the number of occurrence of a word matter in our prediction which seem to be relevant in classifying apps into categories. For example, if the word “fun” occurs multiple times in the Games category and a very few times in the Education category, it shows that the word “fun” is more important for the Game category and less important for the Education category. Alternatively, the Bernoulli Naïve Bayes classifier is used when the absence of a particular word matters. For example, if the word “fun” does not occur in the Business category it is assumed that “fun” cannot be used to classify an app into the Business category because it does not exist there. The Bernoulli Naïve Bayes classifier is commonly used in classifying Spam and Adult contents.

In this research, we lay emphasis on the Multinomial Naïve Bayes classifier because the number of occurrence of a word in an app detail is important in classifying the app into a category. The Multinomial Naïve Bayes classification as described by \cite{Manning2008} is represented as (\ref{eq:eq5}).
\begin{equation}\label{eq:eq5}
P(t|c) = \frac{T_{ct}}{\sum\limits_{t^\prime \in V}T_{ct^\prime}}
\end{equation}

\begin{figure}[h]
\centering
\includegraphics[width=0.5\textwidth]{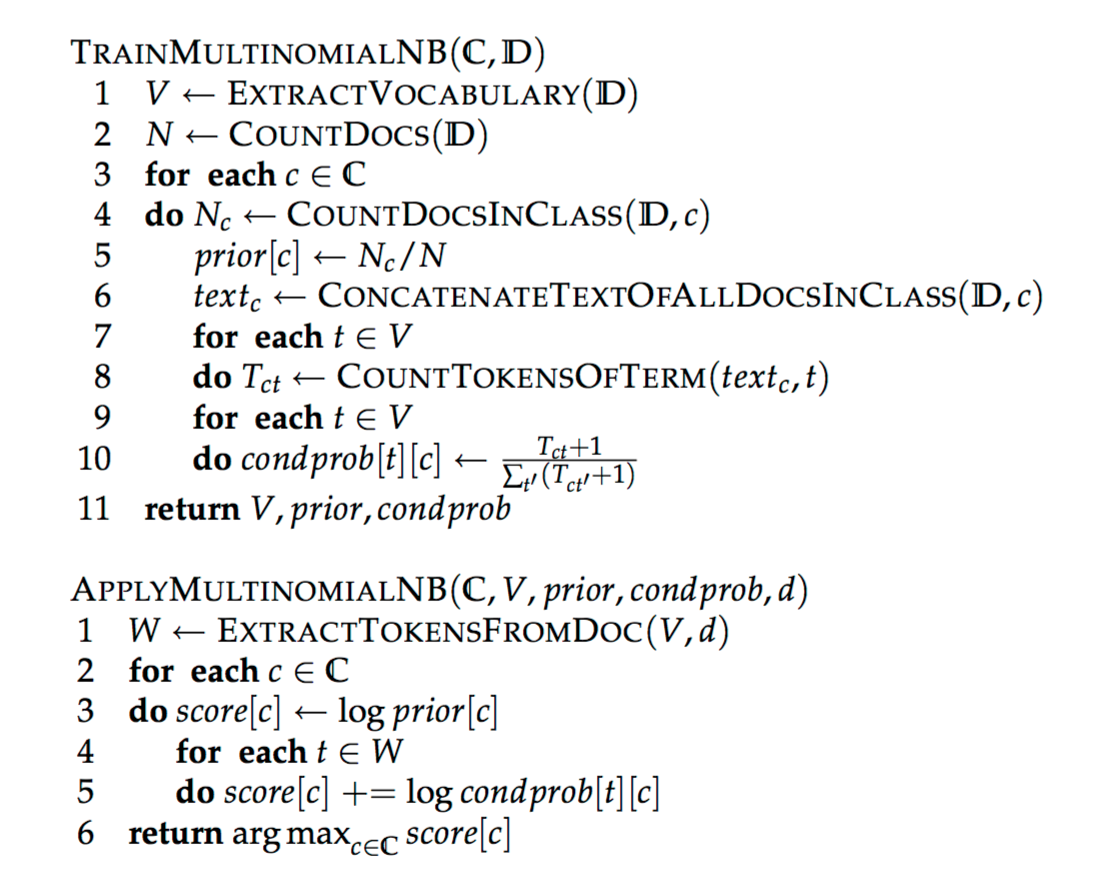}
\caption[Caption for MNB]{Multinomial Naïve Bayes Algorithm \footnotemark}
\label{fig:fig1}
\end{figure}

This estimates the probability of a term $t$ given the category $c$ as the relative frequency of term $t$ in apps belonging to category $c$. The algorithm to be used as described in \cite{Manning2008} is shown in Figure \ref{fig:fig1}

\footnotetext{Source: http://nlp.stanford.edu/IR-book/html/htmledition/naive-bayes-text-classification-1.html}

\begin{figure}[h]
\centering
\includegraphics[width=0.5\textwidth]{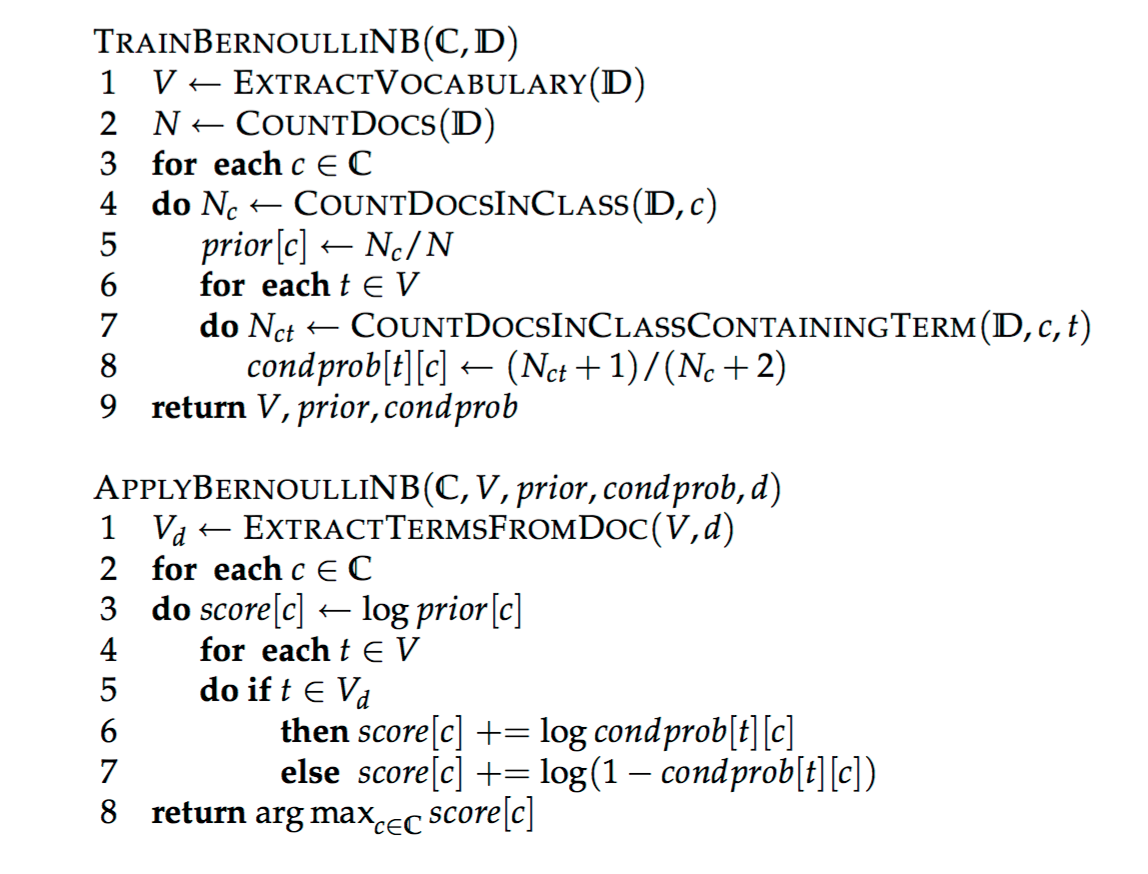}
\caption[Caption for BNB]{Bernoulli Naïve Bayes Algorithm \footnotemark}
\label{fig:fig2}
\end{figure}

The Bernoulli Naïve Bayes classifier differs from Multinomial Naive Bayes classifier as it does not take into account the number of occurrence of the word. Bernoulli Naïve Bayes provides a Boolean indicator for the occurrence of a word as 1 and 0 if the word does not exist. The algorithm that will be used in this project as described in  \cite{Manning2008} is shown in Figure \ref{fig:fig2}.

\footnotetext{Source: http://nlp.stanford.edu/IR-book/html/htmledition/the-bernoulli-model-1.html}

\section{Dataset Overview}
The dataset used in this project is the metadata of 1,197,995 of 1,390,545 apps after filtering out bad data. This dataset is a CSV file containing apps extracted from Google Play Store as at June 2015 with GooglePlayStoreCrawler\footnote{GooglePlayStoreCrawler:\\ https://github.com/MarcelloLins/GooglePlayAppsCrawler}. This data contains the following attributes:

\begin{center}
\begingroup
    \fontsize{8pt}{9pt}\selectfont
\texttt{AppName, Developer, IsTopDeveloper, DeveloperURL, DeveloperNormalizedDomain, Category, IsFree, Price, Reviewers, Score.Total, Score.Count, Score.FiveStars, Score.FourStars, Score.ThreeStars, Score.TwoStars, Score.OneStars, Instalations, CurrentVersion, MinimumOSVersion, ContentRating, HaveInAppPurchases, DeveloperEmail, DeveloperWebsite, PhysicalAddress, LastUpdateDate, Description\\
}
\endgroup
\end{center}

The data contains 41 categories out of the current 43 categories available on Google Play, leaving out “wallpapers” and “widget” which were formally classified as “personalization” before being separated later. To ensure the correctness of our data, we extracted all apps uploaded by top developers, leaving us with 10,369 apps, labelled by their category. These apps will be used to train and test our classifier because they provide a verified source for classification. Providing good data for the classifier will improve its accuracy. Furthermore, it is very likely that successful developers will take much time to properly categorise their apps, provide eloquent descriptions for their apps and ensure their details are complete.

\subsection{Feature Processing}
To create an accurate predictive model, we selected five attributes from all applications; this will be used to extract our features for each category later. These attributes include:
\begin{itemize}
  \item AppName
  \item ContentRating
  \item IsFree 
  \item HaveInAppPurchases 
  \item Description\\
\end{itemize}

Using the bag-of-words model, we extracted tokens from our 10,369 apps after combining these attributes together, removing stop words such as “the, as, is, who, on, …”, removing numbers, punctuation and setting all words to lowercase. At this stage, we have 113,463 features that were extracted from 10,369 app details using Term Frequency-Inverse Document Frequency (TF-IDF) statistics. TF-IDF allows us to reduce the impact of tokens that occurs frequently, so that they do not affect features that occur in small amounts. The formula used to calculate TF-IDF is given as (\ref{eq:eq6}) as described in \cite{Eck2005}.

\begin{equation}\label{eq:eq6}
w_n = TF_n \times log(IDF_n) 
\end{equation}

Where $w_n$ represents a word in the vector $(w_0, w_1, w_2, ..., w_n)$ for each app, represented as a document $D$.

$TF_n$ is the Term Frequency of the n-th word in document $D$.

$IDF_n$ is represented as the Inverse Document Frequency of the n-th term in document $D$ represented as:
\begin{align*}
\frac{\# documents}{\# documents\ containing\ the\ n-th\ term}\\
\end{align*}

\begin{table}[]
\centering
\caption{Aggregate distribution of top developers dataset}
\label{tab:tab1}
\begin{tabular}{@{}lcl@{}}
\toprule
\multicolumn{1}{c}{\textbf{Category}} & \textbf{No of Apps} & \multicolumn{1}{c}{\textbf{Top 3 words}} \\ \midrule
BOOKS\_AND\_REFERENCE                 & 427                 & quickly collins feature                  \\
BUSINESS                              & 158                 & dynamics android files                   \\
COMICS                                & 19                  & dc aitype spiderman                      \\
COMMUNICATION                         & 135                 & send voice chat                          \\
EDUCATION                             & 905                 & offline kids students                    \\
ENTERTAINMENT                         & 407                 & live disney episodes                     \\
FINANCE                               & 136                 & payments pay account                     \\
GAME\_ACTION                          & 440                 & world jump use                           \\
GAME\_ADVENTURE                       & 276                 & minigames play object                    \\
GAME\_ARCADE                          & 641                 & world games mode                         \\
GAME\_BOARD                           & 158                 & board hidden masterthis                  \\
GAME\_CARD                            & 158                 & reel machines poker                      \\
GAME\_CASINO                          & 94                  & vegas coins game                         \\
GAME\_CASUAL                          & 1006                & hints fun modes                          \\
GAME\_EDUCATIONAL                     & 186                 & math educational learn                   \\
GAME\_MUSIC                           & 26                  & cinderella hits rhythm                   \\
GAME\_PUZZLE                          & 808                 & match swap play                          \\
GAME\_RACING                          & 174                 & racer bike driving                       \\
GAME\_ROLE\_PLAYING                   & 291                 & play characters story                    \\
GAME\_SIMULATION                      & 192                 & simulator make city                      \\
GAME\_SPORTS                          & 261                 & teams flick ball                         \\
GAME\_STRATEGY                        & 311                 & build tower play                         \\
GAME\_TRIVIA                          & 97                  & play avatar new                          \\
GAME\_WORD                            & 55                  & phrase challenge guess                   \\
HEALTH\_AND\_FITNESS                  & 124                 & sleep health training                    \\
LIBRARIES\_AND\_DEMO                  & 52                  & effects thirdparty aviarys               \\
LIFESTYLE                             & 184                 & alarm application cooking                \\
MEDIA\_AND\_VIDEO                     & 87                  & la torrent movies                        \\
MEDICAL                               & 95                  & feature pregnancy drug                   \\
MUSIC\_AND\_AUDIO                     & 124                 & bass tracks sound                        \\
NEWS\_AND\_MAGAZINES                  & 408                 & breaking subscribers read                \\
PERSONALIZATION                       & 242                 & battery use launcher                     \\
PHOTOGRAPHY                           & 187                 & zoom share instagram                     \\
PRODUCTIVITY                          & 341                 & appplease qs download                    \\
SHOPPING                              & 124                 & save list store                          \\
SOCIAL                                & 108                 & messages share dating                    \\
SPORTS                                & 139                 & nfl nba scores                           \\
TOOLS                                 & 285                 & battery google use                       \\
TRANSPORTATION                        & 65                  & taxi transit checker                     \\
TRAVEL\_AND\_LOCAL                    & 362                 & plan travel streets                      \\
WEATHER                               & 81                  & force limit temperature                   \\ \bottomrule
\end{tabular}
\end{table}

The TF-IDF method is more preferable than a regular frequency count of tokens \cite{Baeza-Yates1999}; this is because specific words are used to describe specific categories in the store, such as “fun” in “Games” category and “money” in “Business” category. The number of occurrence of these words determine how important they are for all apps. Table \ref{tab:tab1} shows the aggregate distributions for each app category in our 10,369 app dataset. 

\subsection{Further Processing}
Further processing was done to the 10,369 app dataset in order to remove attributes or features with little significance. In this process, the TF-IDF method was used again to reduce the number of features from 113,463 to 14,571, by removing words that occur in more than 70\% and words that occur in less than 0.05\% of all 10,369 apps. Furthermore, apps were grouped into two filters as shown below:\\

\begin{enumerate}
\item{\textit{All Apps:} All apps contain all 10,369 apps with 14,571 optimal features after preprocessing.}
\item{\textit{Filtered Apps:} Filtered apps contain 8,366 apps after apps with low description (less than a simple paragraph) were removed. This ensures we remove apps that are not descriptive enough. A descriptive app is about four to six sentences which is approximately 100 words \footnote{Paragraph Length: https://strainindex.wordpress.com/2010/10/25/plain-paragraph-length/}. Furthermore, categories will little support, i.e. categories with very few apps were removed. These categories are (COMICS, LIBRARIES\_AND\_DEMO, GAME\_MUSIC, GAME\_WORD) and they were removed because when the dataset is split into test and training data, the categories will have very few support and thus affect the performance of the classifier.}\\
\end{enumerate}

For each filter above, apps were further grouped into four categories:\\

\begin{enumerate}
\item{\textit{OnlyGameApps:} These are apps with the “GAME\_” category. In “Filtered Apps” there are 4,374 games with 15 categories and in “All apps” there are 5,174 games with 17 categories.}
\item{\textit{GroupedGameApps:} These apps are all 10,369 apps where all “GAME\_” categories are grouped together, in this group, all games are grouped as “GAMES” giving us 25 categories other than the 41 initial categories.}
\item{\textit{OnlyOtherCategories:} This contains all apps in other categories except games. With this, we have 24 categories and 5,195 apps in the “All apps” filter.}
\item{\textit{AllCategories:} TThese are all 41 categories from the 10,369 apps and 37 categories from 8,366 filtered apps.}\\
\end{enumerate}

Table \ref{tab:tab2} shows the distribution of the grouped apps used in evaluating the performance of the classifier.

\begin{table}[h]
\centering
\caption{distribution of grouped apps}
\label{tab:tab2}
\begin{tabular}{|c|c|c|c|c|}
\hline
\multirow{2}{*}{\textbf{Filters}} & \multicolumn{2}{c|}{\textbf{All Apps}}     & \multicolumn{2}{c|}{\textbf{Filtered Apps}} \\ \cline{2-5} 
                                  & \textbf{Categories} & \textbf{Apps} & \textbf{Categories}  & \textbf{Apps} \\ \hline
\textit{AllCategories}            & 41                  & 10,369               & 37                   & 8,366                \\ \hline
\textit{OnlyOtherCategories}      & 24                  & 5,195                & 22                   & 3,992                \\ \hline
\textit{GroupedGameApps}          & 25                  & 10,369               & 23                   & 8,366                \\ \hline
\textit{OnlyGameApps}             & 17                  & 5,174                & 15                   & 4,374                \\ \hline
\end{tabular}
\end{table}

\subsection{Prior and Likelihood Formation}
Given our dataset, we can calculate our prior and likelihood for each category using Naïve Bayes as 
\begin{align*}
c_{MAP} &= \argmax_{c\in C} P(app|c) P(c)\\
 &= \argmax_{c\in C}  P(w_0, w_1, w_2,..., w_n|c) P(c)
 \end{align*}
 
 For example, the prior for the "BUSINESS" category is calculated as: 
\begin{align*} 
 P(b) = \frac{N_{b}}{N} = \frac{158}{10369} = 0.015
 \end{align*}
The likelihood using multinomial Naïve Bayes can be calculated as: 
\begin{align*} 
P(w| b) = \frac{count(w,b)}{count(b)},  \forall w\in W
\end{align*} 

From this we can estimate the probability of a category given an app using Naïve Bayes as (\ref{eq:eq7}).
\begin{equation}\label{eq:eq7}
c_{NB} = \argmax_{c\in C} P(c_j) \prod_{w\in W} P(w|c)
\end{equation}\\

\section{Results}
The algorithm was implemented using Python and the  “sklearn” \footnote{Sklearn: http://scikit-learn.org} machine learning library. Sklearn is a machine learning library in Python built on NumPy, SciPy, and matplotlib \footnote{SciPy Libraries: http://scipy.org }. The library allows us to perform data mining and data analysis such as clustering, regression analysis, classification and preprocessing of data. 

\subsection{Evaluating the model}
Evaluating the model used for classifying apps on Google Play involved using various cross-validation methods to test its performance. Other evaluation strategies involved testing the model with the training and testing data, drawing a confusion matrix to determine true positives/negatives and false positive/negatives. Furthermore, a learning curve was plotted to see how the model performs as the training and testing set increases and a Recursive Feature Elimination with cross-validation applied to see how the model performs when eliminating features for each iteration. The following highlights how this classifier performs using these methods with the \textit{AllCategories} apps and \textit{GroupedGameApps} apps as described in the previous section. \textit{AllCategories} represents 41 categories from 10,369 apps, and the \textit{GroupedGameApps} represents 25 categories from 10,369 applications when all games are grouped together as a single category.\\

\subsubsection{\textit{Training and Test Data evaluation}}
\begin{figure}[h]
\centering
\includegraphics[width=0.5\textwidth]{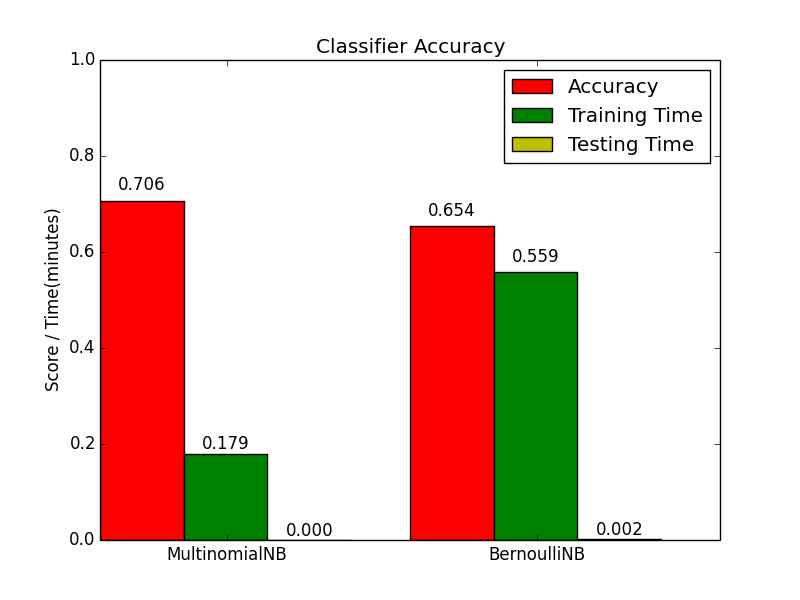}
\caption{\textit{AllCategories} classifier results}
\label{fig:fig3}
\end{figure}

The data was split into training and testing dataset of approximately 80\% to 20\%. The training dataset was used to train the classifier for both Multinomial and Bernoulli classifiers and then used to predict the outcome of the test dataset. Figure \ref{fig:fig3} shows the performance of the classifier for all 10,369 apps.

\begin{figure}[h]
\centering
\includegraphics[width=0.5\textwidth]{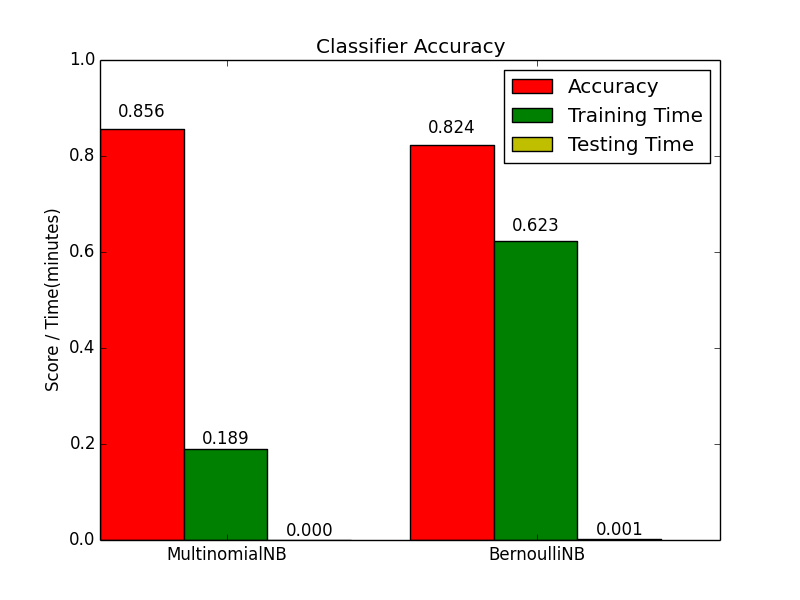}
\caption{\textit{GroupedGameApps} classifier results}
\label{fig:fig4}
\end{figure}

From the results above, it is seen that the Multinomial Naïve Bayes classifier performs better than the Bernoulli Naïve Bayes classifier with about 70\% accuracy. The classifier is also faster in terms of computation speed than the Bernoulli Naïve Bayes classifier. To improve accuracy, all “games” were grouped into a “GAMES” category, reducing the number of categories to 25 and tested with the classifier. Figure \ref{fig:fig4} shows a significant improvement in the results with 85.6\% accuracy in the Multinomial Naïve Bayes and 82.4\% accuracy in the Bernoulli Naïve Bayes.

\subsubsection{K-Fold Cross Validation}
Using K-Fold cross-validation, a 2-Fold and 10-Fold cross-validation for both Multinomial Naïve Bayes and the Bernoulli Naïve Bayes classifiers was computed to determine how each classifier performs overall. Table \ref{tab:tab3} shows the average score for both the 2-Fold and the 10-Fold cross-validation.

\begin{table}[h]
\centering
\caption{Average Score for 2-Fold and 10-Fold Cross validation}
\label{tab:tab3}
\begin{tabular}{|c|c|c|c|c|}
\hline
\multirow{2}{*}{\textbf{Filters}} & \multicolumn{2}{c|}{\textbf{\begin{tabular}[c]{@{}c@{}}Multinomial \\ Score\end{tabular}}} & \multicolumn{2}{c|}{\textbf{\begin{tabular}[c]{@{}c@{}}Bernoulli \\ Score\end{tabular}}} \\ \cline{2-5} 
                                  & \textbf{2-Fold}                             & \textbf{10-Fold}                             & \textbf{2-Fold}                            & \textbf{10-Fold}                            \\ \hline
\textit{AllCategories}            & 0.66                                        & 0.689                                        & 0.623                                      & 0.65                                        \\ \hline
\textit{GroupedGameApps}          & 0.836                                       & 0.849                                        & 0.805                                      & 0.819                                       \\ \hline
\end{tabular}
\end{table}

From the results above, it can be seen that the Multinomial Naïve Bayes classifier still performs better than the Bernoulli. Also, the average classifier score improves as the number of folds’ increases.\\

\subsubsection{Recursive Feature Elimination using StratifiedKFold Cross Validation}
\begin{figure}[h]
\centering
\includegraphics[width=0.5\textwidth]{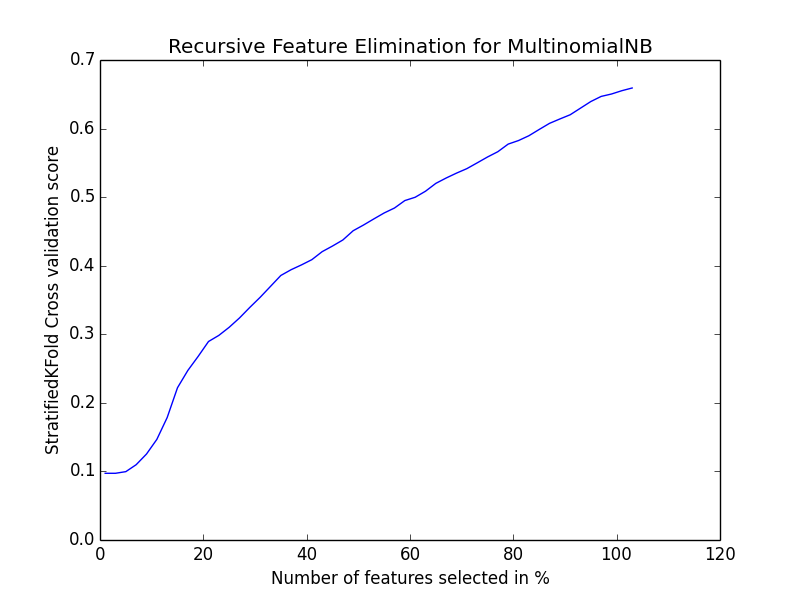}
\caption{RFE for \textit{AllCategories}}
\label{fig:fig5}
\end{figure}

Recursive Feature Elimination involves selecting features recursively and then reducing the number of features. Here, the classifier is trained with the initial set of features and scored. Then, features are gradually pruned from the current set of features in each iteration and scored. In this test, the StratifiedKFold cross-validation involving splitting the data n-times while shuffling the dataset was used with a Recursive Feature Elimination. Stratification involves rearranging the data as to ensure each fold is a good representative of the whole. In this test, 20\% of the features were removed for each iteration to determine the number of features that performs best with the model. Figure \ref{fig:fig5} shows a Recursive Feature Elimination (RFE) with cross-validation for the Multinomial Naïve Bayes Classifier for all categories.

\begin{figure}[h]
\centering
\includegraphics[width=0.5\textwidth]{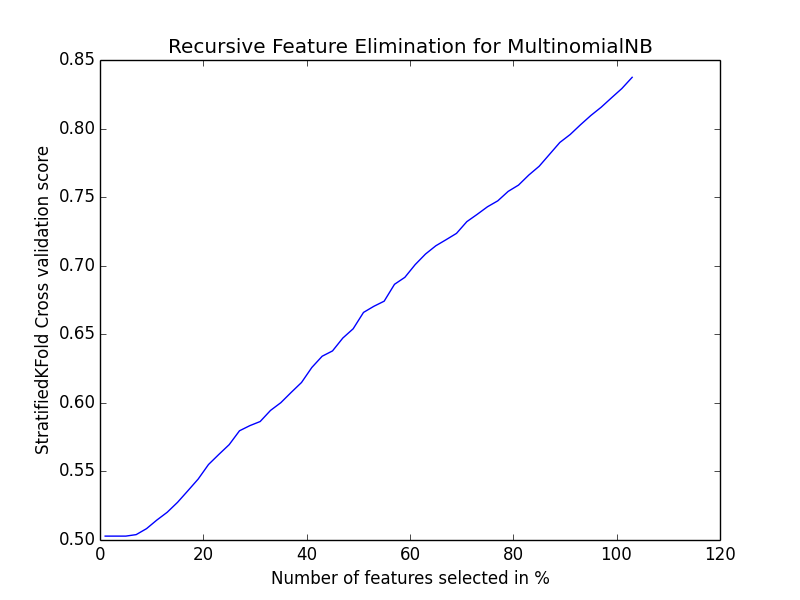}
\caption{RFE for \textit{GroupedGameApps}}
\label{fig:fig6}
\end{figure}

In Figure \ref{fig:fig6}, we can see a significant improvement in the result when all games are grouped together. The classifier performs well initially with about 20\% of the features used resulting in a 50\% accuracy compared to the 10\% accuracy observed when all 41 categories are used in Figure \ref{fig:fig5}.\\

\subsubsection{Learning Curve}
\begin{figure}[h]
\centering
\includegraphics[width=0.5\textwidth]{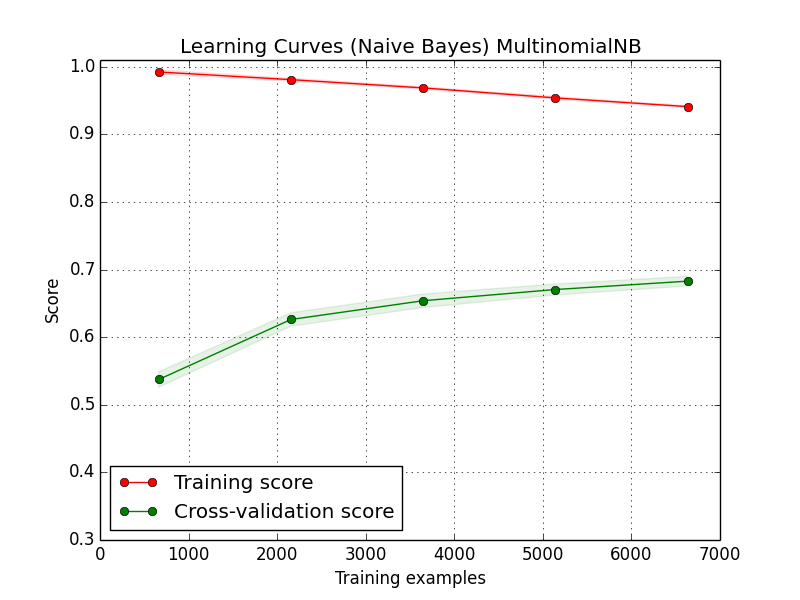}
\caption{Learning curve for \textit{AllCategories}}
\label{fig:fig7}
\end{figure}

A learning curve is a graphical representation of the increase in learning with experience. The experience of the classifier increases as the number of training dataset increases thus the precision or accuracy of the algorithm is improved. Figure \ref{fig:fig7} shows the learning curves for Multinomial Naïve Bayes for the 41 categories (\textit{AllCategories}).  

\begin{figure}[h]
\centering
\includegraphics[width=0.5\textwidth]{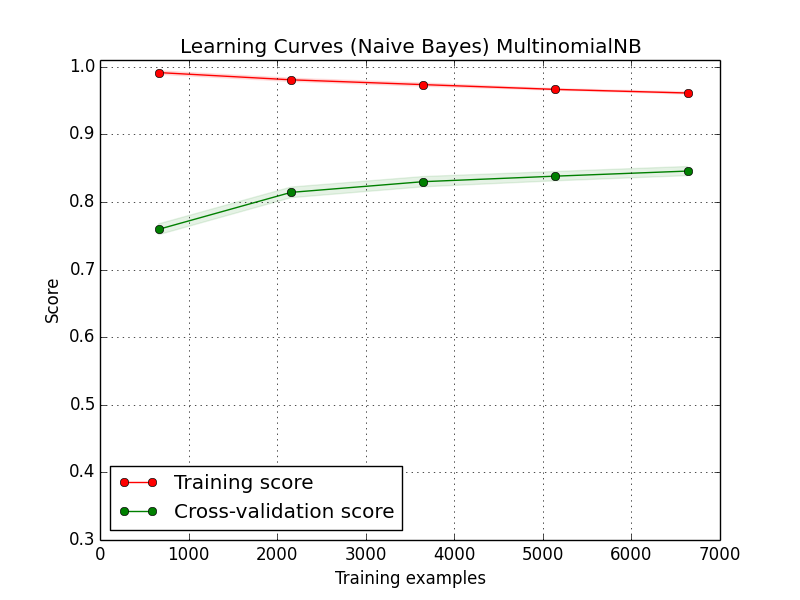}
\caption{Learning curve for \textit{GroupedGameApps}}
\label{fig:fig8}
\end{figure}

Significant improvement is seen in Figure \ref{fig:fig8} when all games are grouped together.\\

\subsubsection{Classification Report}
Table \ref{tab:tab4} shows the general statistics of the Multinomial Naïve Bayes classifier for \textit{AllCategories}. This classifier is selected because it performs better than the Bernoulli Naïve Bayes classifier. In Table \ref{tab:tab4}, we show the precision (accuracy), recall, and f1-score which is the harmonic mean of precision and recall. False Positives (FP), False Negatives (FN), True Positive (TP), True Negative (TN), True Positive Ratio (TPR), False Negative Ratio (FNR) and support for the classifier derived from the confusion matrix for the 2,074 testing dataset is also displayed.

From the results obtained, a confidence level of around 70\% shows that the algorithm performs well in determining the category of a set of applications. However, because the "Game" category has several sub-categories, the algorithm might misclassify some gaming apps. For example, an “Action Game” might be misclassified as an “Arcade Game”, and this will eventually affect the performance of the model because they both have similar words used to describe them. However, grouping the "Games" category significantly improved the performance of the model to about 85\%. We can further determine how gaming apps perform by training the algorithm on the “games” category only and further compare that with other categories apart from games. \\

\subsection{Improving Results}
\begin{figure}[h]
\centering
\includegraphics[width=0.5\textwidth]{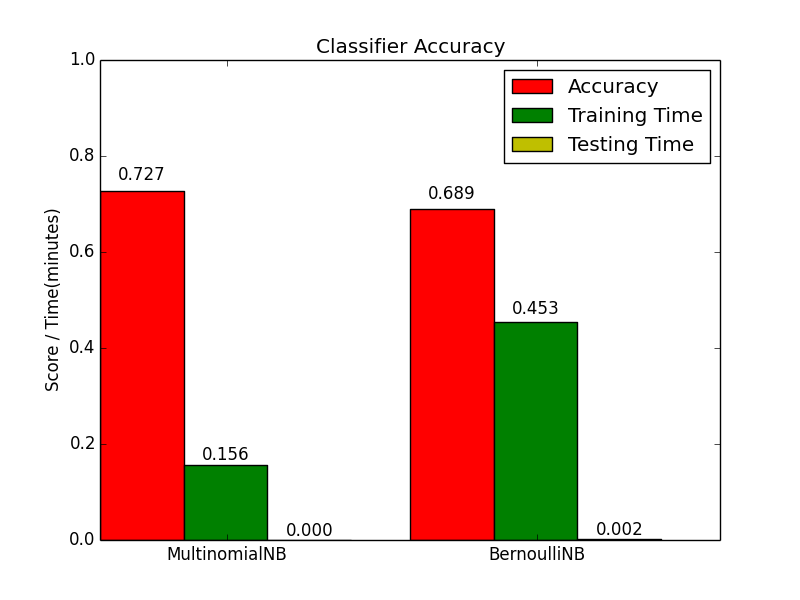}
\caption{Improved classifier results}
\label{fig:fig9}
\end{figure}

The dataset was further filtered to remove categories with few applications and remove apps with few words describing them in order to improve accuracy. The resulting filtered dataset contains 8,366 apps and 37 categories compared to the 10,369 apps used previously. The results show that further optimization of the dataset can significantly improve performance. Figure \ref{fig:fig9} reveals the improvement over the previous result in both Multinomial and Bernoulli Naïve Bayes algorithms for all 37 categories.\\

\begin{figure}[h]
\centering
\includegraphics[width=0.5\textwidth]{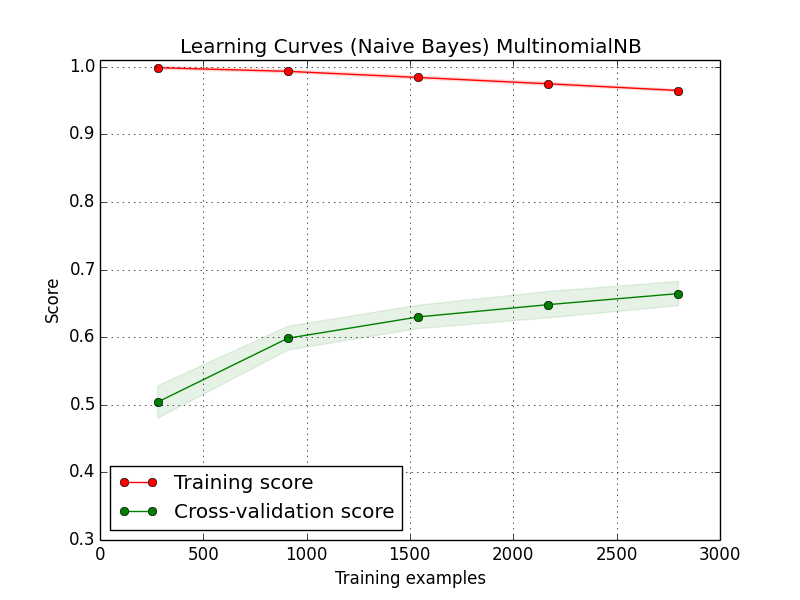}
\caption{Learning curve for all Gaming applications}
\label{fig:fig10}
\end{figure}

Furthermore, grouping all games in a single category, significantly improved the overall performance of the classifier. The Multinomial Naïve Bayes classifier improved to about 87\% accuracy from 85\% accuracy observed in the previous results. We can determine the effect of game categories by plotting a learning curve on all 4,374 games to discover how the classifier performs on all 15 game categories. Figure \ref{fig:fig10} shows the learning curve for all filtered Game categories using 20\% testing dataset from the 4,374 games.\\

\begin{figure}[h]
\centering
\includegraphics[width=0.5\textwidth]{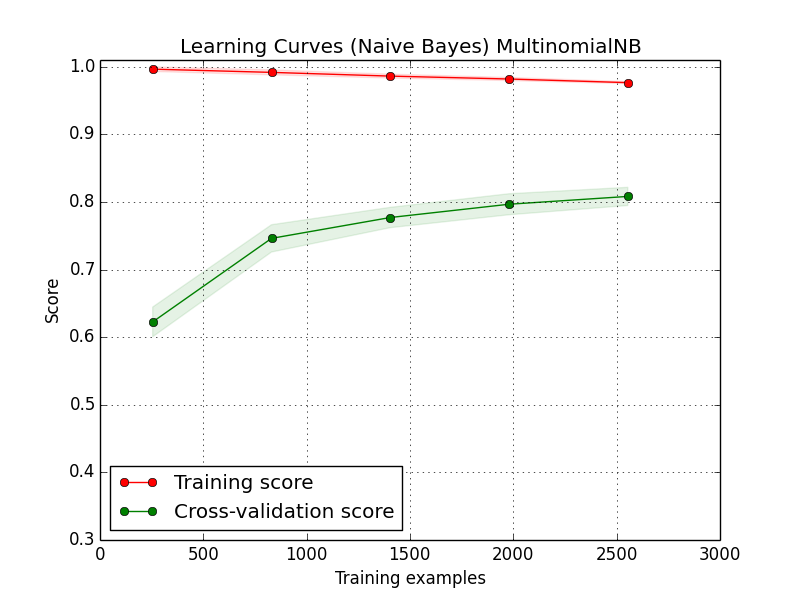}
\caption{Learning curve for other categories with the exemption of games }
\label{fig:fig11}
\end{figure}

From the results, the classifier was 67\% accurate in categorizing games. This result can be compared to all other categories apart from games. Figure \ref{fig:fig11} shows that the classifier performed better when classifying apps in other categories than classifying games. The result observed might be as a result of close similarity in the words chosen to describe a game; it is also possible that games with similar features will be described with similar words even if they are in different categories resulting in false negatives.

\clearpage
\onecolumn
\begin{table}[]
\centering
\caption{Classification report for 2,074 Test dataset with Multinomial Naïve Bayes}
\label{tab:tab4}
\begin{tabular}{|c|c|c|c|c|c|c|c|c|c|c|}
\hline
\textbf{Category}     & \textbf{TP} & \textbf{TN} & \textbf{FP} & \textbf{FN} & \textbf{TPR} & \textbf{FNR} & \textbf{precision} & \textbf{recall} & \textbf{f1-score} & \textbf{support} \\ \hline
BOOKS\_AND\_REFERENCE & 76          & 1988        & 1           & 9           & 0.89         & 0.11         & 0.99               & 0.89            & 0.94              & 85               \\ \hline
BUSINESS              & 22          & 2017        & 19          & 16          & 0.58         & 0.42         & 0.54               & 0.58            & 0.56              & 38               \\ \hline
COMICS                & 1           & 2068        & 3           & 2           & 0.33         & 0.67         & 0.25               & 0.33            & 0.29              & 3                \\ \hline
COMMUNICATION         & 21          & 2039        & 6           & 8           & 0.72         & 0.28         & 0.78               & 0.72            & 0.75              & 29               \\ \hline
EDUCATION             & 175         & 1844        & 18          & 37          & 0.83         & 0.17         & 0.91               & 0.83            & 0.86              & 212              \\ \hline
ENTERTAINMENT         & 45          & 1964        & 30          & 35          & 0.56         & 0.44         & 0.6                & 0.56            & 0.58              & 80               \\ \hline
FINANCE               & 16          & 2046        & 9           & 3           & 0.84         & 0.16         & 0.64               & 0.84            & 0.73              & 19               \\ \hline
GAME\_ACTION          & 41          & 1931        & 51          & 51          & 0.45         & 0.55         & 0.45               & 0.45            & 0.45              & 92               \\ \hline
GAME\_ADVENTURE       & 24          & 2010        & 20          & 20          & 0.55         & 0.45         & 0.55               & 0.55            & 0.55              & 44               \\ \hline
GAME\_ARCADE          & 79          & 1871        & 77          & 47          & 0.63         & 0.37         & 0.51               & 0.63            & 0.56              & 126              \\ \hline
GAME\_BOARD           & 24          & 2031        & 7           & 12          & 0.67         & 0.33         & 0.77               & 0.67            & 0.72              & 36               \\ \hline
GAME\_CARD            & 27          & 2029        & 5           & 13          & 0.68         & 0.33         & 0.84               & 0.68            & 0.75              & 40               \\ \hline
GAME\_CASINO          & 19          & 2048        & 3           & 4           & 0.83         & 0.17         & 0.86               & 0.83            & 0.84              & 23               \\ \hline
GAME\_CASUAL          & 145         & 1785        & 90          & 54          & 0.73         & 0.27         & 0.62               & 0.73            & 0.67              & 199              \\ \hline
GAME\_EDUCATIONAL     & 209         & 1810        & 23          & 32          & 0.87         & 0.13         & 0.56               & 0.76            & 0.65              & 29               \\ \hline
GAME\_MUSIC           & 0           & 2067        & 1           & 6           & 0.00         & 1.00         & 0                  & 0               & 0                 & 6                \\ \hline
GAME\_PUZZLE          & 88          & 1878        & 53          & 55          & 0.62         & 0.38         & 0.62               & 0.62            & 0.62              & 143              \\ \hline
GAME\_RACING          & 21          & 2036        & 7           & 10          & 0.68         & 0.32         & 0.75               & 0.68            & 0.71              & 31               \\ \hline
GAME\_ROLE\_PLAYING   & 47          & 1983        & 29          & 15          & 0.76         & 0.24         & 0.62               & 0.76            & 0.68              & 62               \\ \hline
GAME\_SIMULATION      & 16          & 2025        & 12          & 21          & 0.43         & 0.57         & 0.57               & 0.43            & 0.49              & 37               \\ \hline
GAME\_SPORTS          & 63          & 1983        & 11          & 17          & 0.79         & 0.21         & 0.83               & 0.76            & 0.8               & 51               \\ \hline
GAME\_STRATEGY        & 52          & 1997        & 14          & 11          & 0.83         & 0.17         & 0.79               & 0.83            & 0.81              & 63               \\ \hline
GAME\_TRIVIA          & 11          & 2057        & 3           & 3           & 0.79         & 0.21         & 0.79               & 0.79            & 0.79              & 14               \\ \hline
GAME\_WORD            & 3           & 2063        & 4           & 4           & 0.43         & 0.57         & 0.43               & 0.43            & 0.43              & 7                \\ \hline
HEALTH\_AND\_FITNESS  & 21          & 2050        & 0           & 3           & 0.88         & 0.13         & 1                  & 0.88            & 0.93              & 24               \\ \hline
LIBRARIES\_AND\_DEMO  & 2           & 2067        & 0           & 5           & 0.29         & 0.71         & 1                  & 0.29            & 0.44              & 7                \\ \hline
LIFESTYLE             & 15          & 2021        & 18          & 20          & 0.43         & 0.57         & 0.45               & 0.43            & 0.44              & 35               \\ \hline
MEDIA\_AND\_VIDEO     & 7           & 2054        & 5           & 8           & 0.47         & 0.53         & 0.58               & 0.47            & 0.52              & 15               \\ \hline
MEDICAL               & 13          & 2056        & 0           & 5           & 0.72         & 0.28         & 1                  & 0.72            & 0.84              & 18               \\ \hline
MUSIC\_AND\_AUDIO     & 28          & 2038        & 4           & 4           & 0.88         & 0.13         & 0.88               & 0.88            & 0.88              & 32               \\ \hline
NEWS\_AND\_MAGAZINES  & 74          & 1963        & 20          & 17          & 0.81         & 0.19         & 0.79               & 0.81            & 0.8               & 91               \\ \hline
PERSONALIZATION       & 35          & 2022        & 8           & 9           & 0.80         & 0.20         & 0.81               & 0.8             & 0.8               & 44               \\ \hline
PHOTOGRAPHY           & 32          & 2031        & 7           & 4           & 0.89         & 0.11         & 0.82               & 0.89            & 0.85              & 36               \\ \hline
PRODUCTIVITY          & 47          & 1991        & 15          & 21          & 0.69         & 0.31         & 0.76               & 0.69            & 0.72              & 68               \\ \hline
SHOPPING              & 16          & 2038        & 8           & 12          & 0.57         & 0.43         & 0.67               & 0.57            & 0.62              & 28               \\ \hline
SOCIAL                & 16          & 2047        & 2           & 9           & 0.64         & 0.36         & 0.89               & 0.64            & 0.74              & 25               \\ \hline
SPORTS                & 22          & 2040        & 5           & 7           & 0.76         & 0.24         & 0.81               & 0.76            & 0.79              & 29               \\ \hline
TOOLS                 & 27          & 2010        & 22          & 15          & 0.64         & 0.36         & 0.55               & 0.64            & 0.59              & 42               \\ \hline
TRANSPORTATION        & 13          & 2058        & 0           & 3           & 0.81         & 0.19         & 1                  & 0.81            & 0.9               & 16               \\ \hline
TRAVEL\_AND\_LOCAL    & 69          & 1985        & 8           & 12          & 0.85         & 0.15         & 0.9                & 0.85            & 0.87              & 81               \\ \hline
WEATHER               & 13          & 2059        & 1           & 1           & 0.93         & 0.07         & 0.93               & 0.93            & 0.93              & 14               \\ \hline
                      &             &             &             &             &              &              &                    &                 &                   &                  \\ \hline
\textbf{Total}        &             &             &             &             &              &              &                    &                 &                   & \textbf{2074}    \\ \hline
\end{tabular}
\end{table}

\begin{table}[]
\centering
\caption{Confusion Matrix for game categories with Multinomial Naïve Bayes classification}
\label{tab:tab5}
\begin{tabular}{|c|c|c|c|c|c|c|c|c|c|c|c|c|c|c|c|c|}
\hline
\textbf{Key} & \textbf{Category}            & \textbf{1} & \textbf{2} & \textbf{3} & \textbf{4} & \textbf{5} & \textbf{6} & \textbf{7} & \textbf{8} & \textbf{9} & \textbf{10} & \textbf{11} & \textbf{12} & \textbf{13} & \textbf{14} & \textbf{15} \\ \hline
\textbf{1}   & \textbf{GAME\_ACTION}        & 41         & 2          & 19         & 0          & 0          & 0          & 6          & 0          & 3          & 2           & 5           & 1           & 2           & 4           & 0           \\ \hline
\textbf{2}   & \textbf{GAME\_ADVENTURE}     & 3          & 21         & 2          & 0          & 0          & 0          & 7          & 1          & 3          & 0           & 4           & 0           & 0           & 0           & 0           \\ \hline
\textbf{3}   & \textbf{GAME\_ARCADE}        & 12         & 0          & 56         & 0          & 0          & 1          & 5          & 0          & 9          & 0           & 6           & 0           & 2           & 5           & 1           \\ \hline
\textbf{4}   & \textbf{GAME\_BOARD}         & 0          & 0          & 1          & 20         & 1          & 0          & 0          & 0          & 5          & 0           & 0           & 0           & 0           & 0           & 0           \\ \hline
\textbf{5}   & \textbf{GAME\_CARD}          & 0          & 0          & 0          & 0          & 18         & 3          & 2          & 0          & 1          & 0           & 2           & 0           & 0           & 0           & 0           \\ \hline
\textbf{6}   & \textbf{GAME\_CASINO}        & 0          & 0          & 0          & 0          & 0          & 16         & 1          & 0          & 0          & 0           & 0           & 0           & 0           & 0           & 0           \\ \hline
\textbf{7}   & \textbf{GAME\_CASUAL}        & 7          & 9          & 14         & 1          & 0          & 0          & 126        & 1          & 11         & 0           & 1           & 4           & 0           & 0           & 0           \\ \hline
\textbf{8}   & \textbf{GAME\_EDUCATIONAL}   & 0          & 0          & 0          & 0          & 0          & 0          & 3          & 27         & 0          & 0           & 0           & 0           & 0           & 1           & 1           \\ \hline
\textbf{9}   & \textbf{GAME\_PUZZLE}        & 4          & 4          & 9          & 2          & 0          & 0          & 17         & 4          & 87         & 0           & 1           & 0           & 0           & 2           & 1           \\ \hline
\textbf{10}  & \textbf{GAME\_RACING}        & 0          & 0          & 2          & 0          & 0          & 0          & 1          & 0          & 0          & 28          & 1           & 0           & 0           & 0           & 0           \\ \hline
\textbf{11}  & \textbf{GAME\_ROLE\_PLAYING} & 5          & 0          & 1          & 0          & 0          & 0          & 1          & 0          & 2          & 0           & 38          & 1           & 0           & 4           & 0           \\ \hline
\textbf{12}  & \textbf{GAME\_SIMULATION}    & 2          & 1          & 1          & 0          & 0          & 0          & 6          & 1          & 1          & 0           & 1           & 22          & 0           & 2           & 0           \\ \hline
\textbf{13}  & \textbf{GAME\_SPORTS}        & 1          & 0          & 4          & 0          & 0          & 0          & 2          & 0          & 0          & 0           & 0           & 1           & 29          & 0           & 0           \\ \hline
\textbf{14}  & \textbf{GAME\_STRATEGY}      & 5          & 3          & 4          & 0          & 0          & 0          & 5          & 0          & 0          & 0           & 0           & 2           & 1           & 43          & 0           \\ \hline
\textbf{15}  & \textbf{GAME\_TRIVIA}        & 0          & 0          & 1          & 0          & 0          & 0          & 3          & 0          & 1          & 0           & 0           & 0           & 3           & 0           & 16          \\ \hline
\end{tabular}
\end{table}
\twocolumn

We can observe the misclassification of games from the confusion matrix in Table \ref{tab:tab5}. The results show 875 games tested against 3,499 games used to train the classifier. From the result, several misclassification is seen in GAME\_ACTION and GAME\_ARCADE, GAME\_CASUAL and GAME\_ARCADE, GAME\_PUZZLE and GAME\_CASUAL, GAME\_ADVENTURE and GAME\_CASUAL. It can also be observed that many apps in these categories are described with similar words from the word tokens thereby resulting in misclassification of those apps. Overall, the algorithm was successful in classifying 67\% of games and 72.7\% of all applications in all categories. Further improvement is seen when all games are grouped and one with 87\% accuracy.

Table \ref{tab:tab6} shows the overall performance of both the Multinomial Naïve Bayes classifier and the Bernoulli Naive Bayes classifier in categorizing Android applications with various filters in our dataset. The table shows the classifier precision on a training and testing dataset of 80\% and 20\% respectively.

\begin{table}[h]
\centering
\caption{Overall Performance of the Classifiers}
\label{tab:tab6}
\begin{tabular}{|c|c|c|c|c|}
\hline
\multirow{2}{*}{\textbf{Filters}} & \multicolumn{2}{c|}{\textbf{All Apps}}    & \multicolumn{2}{c|}{\textbf{Filtered Apps}} \\ \cline{2-5} 
                                  & \textbf{Multi} & \textbf{Ber} & \textbf{Multi}  & \textbf{Ber}  \\ \hline
\textit{AllCategories}            & 0.706                & 0.654              & 0.727                 & 0.689               \\ \hline
\textit{OnlyOtherCategories}      & 0.772                & 0.721              & 0.814                 & 0.78                \\ \hline
\textit{GroupedGameApps}          & 0.856                & 0.824              & 0.871                 & 0.85                \\ \hline
\textit{OnlyGameApps}             & 0.65                 & 0.647              & 0.672                 & 0.67                \\ \hline
\end{tabular}
\end{table}

\section{Discussion}
The results show that Multinomial Naïve Bayes performs better than the Bernoulli Naïve Bayes algorithm; this because the number of occurrence of a word matters a lot in our classification problem.  In Bernoulli Naïve Bayes, the absence of a word matters, and although the results are not as efficient as the Multinomial Naïve Bayes, it can play a role in classifying other problems such as spam detection in emails. The overall result shows that the algorithm is weak at classifying games, because of the similarity in words used to describe most gaming applications in different categories. It can also be observed that the classifier performs better when classifying gaming apps in entirely different categories than others. For example, TRANSPORTATION, MEDICAL and WEATHER categories all have a precision of 100\%. Words such as “raining” in WEATHER will most likely not occur in many other categories, same as “doctor” or “pregnancy” in the MEDICAL category. Furthermore, the dataset used do not necessarily mean that the applications were classified accurately. Top developers on Google Play might misclassify some of their applications and eventually affect the performance of the classifier. Alternatively, misclassified applications, especially in the GAMES category may not necessarily mean the app was classified wrongly; this can be observed in “Arcade” and “Action” games as they are sometimes misclassified. 

In general, Naïve Bayes is useful when it comes to text classification because of its speed and performance even with a limited training set. Additionally, this makes Naïve Bayes useful as a baseline for machine learning algorithms in various research. Machine learning plays a significant role in data mining because the computer can learn from past experiences (training), making it useful in analyzing large dataset difficult for humans to comprehend. From data acquisition to data optimization and algorithm selection, we can see that proper data collection and optimization can significantly improve the overall performance of the classifier. Furthermore, selecting the right algorithm for a classification problem is also important, just as the Multinomial Naive Bayes algorithm performed better than the Bernoulli Naïve Bayes algorithm in our classification problem. Additionally, overfitting can be reduced as seen from the learning curve when more data is used to train the classifier.

\section{Future Work}
Naïve Bayes assumes features are independent, and this is likely not the case in many scenarios. There are still several aspects that require improvement, and it is important to note that Naïve Bayes is relatively efficient in document classification as applied in our app categorization, but this can further be improved by using more advanced algorithms. Further enhancements can be made to our classifier as described below. 

\subsection{N-gram usage}
Using 2-grams for example, “alarm clock”, “angry birds” and “music box”, may improve results as these phrases specifically describe a feature. Several n-grams can be combined such as a unigram, 2-grams or 3-grams, to see how precision can be improved with phrases and single words using natural language processing techniques.

\subsection{Other Machine Learning algorithms}
Advanced classification algorithms such as Support Vector Machines (SVM), Hidden Markov Models (HMM) and Decision Trees can further improve categorization accuracy. Although computationally intensive, these algorithms may perform better than the Naïve Bayes classifier.

\section{Summary}
Similar words used to describe various applications in different categories can affect the performance of the classifier. This situation is observed in the gaming categories. Although the error rates are higher than classifying apps in other categories, it can be seen that most misclassification that occur in the “games” category, does not necessarily mean those apps do not belong to the category predicted. For example, an app published in an “Arcade Game” category may also be suitable in an “Action Game” category depending on the type of game. Furthermore, the training data used is not fundamentally a full proof of how apps are classified. Additionally, it can be observed that Multinomial Naïve Bayes performed better than the Bernoulli Naive Bayes classifier in text classification where the number of occurrence of a word is important. In general, proper data collection, optimization and large training set can significantly improve the performance of a machine learning classification algorithm.

\bibliographystyle{IEEEtran}
\bibliography{references}

\end{document}